\documentclass[11pt,a4paper]{article}
\usepackage[hyperref]{emnlp2020}
\usepackage{times}
\usepackage{amsmath}
\usepackage{pgfplots}
\usepackage{latexsym}
\usepackage{url}
\expandafter\def\expandafter\UrlBreaks\expandafter{\UrlBreaks%  save the current one
  \do\a\do\b\do\c\do\d\do\e\do\f\do\g\do\h\do\i\do\j%
  \do\k\do\l\do\m\do\n\do\o\do\p\do\q\do\r\do\s\do\t%
  \do\u\do\v\do\w\do\x\do\y\do\z\do\A\do\B\do\C\do\D%
  \do\E\do\F\do\G\do\H\do\I\do\J\do\K\do\L\do\M\do\N%
  \do\O\do\P\do\Q\do\R\do\S\do\T\do\U\do\V\do\W\do\X%
  \do\Y\do\Z}

\usepackage{xspace}
\usepackage{amsfonts}
\usepackage{array,multirow}
\usepackage{pgfplots}
\usepackage{tikz}
\usepackage{verbatim}
\usepackage{subfig}
\usepackage{arydshln}
\usepackage{booktabs}
\definecolor{ugreen}{rgb}{0,0.5,0}
\definecolor{mygreen}{RGB}{46,139,87}
\definecolor{iyellow}{RGB}{255,250,205}
\definecolor{ipurple}{RGB}{230,230,250}
\definecolor{myred}{RGB}{238,44,44}
\definecolor{myblue}{RGB}{30,144,255}
\definecolor{myorange}{RGB}{255,127,80}
\definecolor{mypurple}{RGB}{255,20,147}

\newcommand{\PreserveBackslash}[1]{\let\temp=\\#1\let\\=\temp}
\newcolumntype{C}[1]{>{\PreserveBackslash\centering}p{#1}}
\newcolumntype{R}[1]{>{\PreserveBackslash\raggedleft}p{#1}}
\newcolumntype{L}[1]{>{\PreserveBackslash\raggedright}p{#1}}
\newcolumntype{M}[1]{>{\centering\arraybackslash}m{#1}}

\newlength{\vseg}
\newlength{\hseg}
\newlength{\wnode}
\newlength{\hnode}
\usepackage{microtype}

\aclfinalcopy % Uncomment this line for the final submission
\title{Shallow-to-Deep Training for Neural Machine Translation}

\author{
  Bei Li$^1$,
  Ziyang Wang$^1$,
  Hui Liu$^1$,
  Yufan Jiang$^1$, \\
  \textbf{Quan Du$^{1,2}$},
  \textbf{Tong Xiao$^{1,2}$\thanks{\xspace\xspace Corresponding author.}},
  \textbf{Huizhen Wang$^{1,2}$}
  \textbf{and Jingbo Zhu$^{1,2}$}\\
  $^{1}$NLP Lab, School of Computer Science and Engineering\\ 
  Northeastern University, Shenyang, China\\
  $^{2}$NiuTrans Research, Shenyang, China \\
  {\tt
        \{libei\_neu,jiangyufan2018,duquanneu\}@outlook.com 
  }\\
  {\tt
        \{wangziyang,huiliu\}@stumail.neu.edu.cn
  }\\
  {\tt
        \{xiaotong,wanghuizhen,zhujingbo\}@mail.neu.edu.cn
  }
}

\date{}

\begin{document}
\maketitle
\begin{abstract}

  Deep encoders have been proven to be effective in improving neural machine translation (NMT) systems, but training an extremely deep encoder is time consuming. Moreover, why deep models help NMT is an open question. In this paper, we investigate the behavior of a well-tuned deep Transformer system. We find that stacking layers is helpful in improving the representation ability of NMT models and adjacent layers perform similarly. This inspires us to develop a shallow-to-deep training method that learns deep models by stacking shallow models. In this way, we successfully train a Transformer system with a 54-layer encoder. Experimental results on WMT'16 English-German and WMT'14 English-French translation tasks show that it is $1.4$ $\times$ faster than training from scratch, and achieves a BLEU score of $30.33$ and $43.29$ on two tasks. The code is publicly available at \href{https://github.com/libeineu/SDT-Training/}{https://github.com/libeineu/SDT-Training}.
  
\end{abstract}

\section{Introduction}
In recent years, neural models have led to state-of-the-art results in machine translation (MT) \cite{bahdanau2014neural,sutskever2014sequence}. Many of these systems can broadly be characterized as following a multi-layer encoder-decoder neural network design: both the encoder and decoder learn representations of word sequences by a stack of layers \cite{vaswani2017attention,wu2016google,gehring2017convs2s}, building on an interesting line of work in improving such models. The simplest of these increases the model capacity by widening the network, whereas more recent work shows benefits from stacking more layers on the encoder side. For example, for the popular Transformer model \cite{vaswani2017attention}, deep systems have shown promising BLEU improvements by either easing the information flow through the network \cite{bapna-etal-2018-training} or constraining the gradient norm across layers \cite{zhang-etal-2019-improving,xu2019lipschitz,liu2020understanding}. An improved system can even learn a 35-layer encoder, which is $5\times$ deeper than that of vanilla Transformer \cite{wang-etal-2019-learning}.

Although these methods have enabled training deep neural MT (NMT) models, questions remain as to the nature of the problem. The main question here is: \textit{why and how deep networks help in NMT}. Note that previous work evaluates these systems in a black-box manner (i.e., BLEU score). It is thus natural to study how much a deep NMT system is able to learn that is different from the shallow counterpart. Beyond this, training an extremely deep model is expensive although a narrow-and-deep network can speed up training \cite{wang-etal-2019-learning}. For example, it takes us $3\times$ longer time to train the model when we deepen the network from 6 layers to 48 layers. This might prevent us from exploiting deeper models in large-scale systems.

In this paper, we explore why deep architectures work to render learning NMT models more effectively. By investigating the change of the hidden states in different layers, we find that new representations are learned by continually stacking layers on top of the base model. More stacked layers lead to a stronger model of representing the sentence. This particularly makes sense in the deep NMT scenario because it has been proven that deep models can benefit from an enriched representation \cite{wang-etal-2019-learning,wu-etal-2019-depth,wei2004multiscale}.

In addition, the finding here inspires us to develop a simple yet efficient method to train a deep NMT encoder: we train model parameters from shallow to deep, rather than training the entire model from scratch. To stabilize training, we design a sparse linear combination method of connecting lower-level layers to the top. It makes efficient pass of information through the deep network but does not require large memory footprint as in dense networks. We experiment with the method in a state-of-the-art deep Transformer system. Our encoder consists of 48-54 layers, which is almost the deepest Transformer model used in NMT. On WMT En-De and En-Fr tasks, it yields a $1.4\times$ speedup of training, matching the state-of-the-art on the WMT'16 En-De task.

\section{Background}
\label{sec:background}

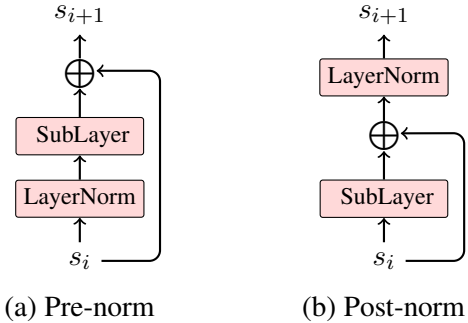
\begin{figure}[t]
	\begin{center}
		\renewcommand{\arraystretch}{0}
		\begin{tabular}{C{.49\textwidth}}
			\subfloat
			{
				\begin{tikzpicture}{baseline}
				\begin{scope}
				
				\setlength{\vseg}{1.5em}
				\setlength{\hseg}{2.5em}
				\setlength{\wnode}{3em}
				\setlength{\hnode}{2em}
				
				\tikzstyle{layernode} = [draw, thin, rounded corners=1pt, inner sep=3pt, fill=yellow!20,  minimum width=0.7\wnode, minimum height=0.3\hnode]
				
				% input
        \node [] (input) at (0,0) {$s_i$};
        \node [] (Pre) at ([yshift=-1em]input.south) {(a) Pre-norm};
				
				% ln
				\node[layernode, fill=red!15] (ln) at ([yshift=\vseg]input.north) {\footnotesize{LayerNorm}};
				
				% module
				\node[layernode,  fill=red!15] (module) at ([yshift=\vseg]ln.north) {\footnotesize{\ \ SubLayer\ \ }};
				
				% residual
				\node[ inner sep=0pt, minimum width=0.15\wnode, minimum height=0.1\hnode] (add1) at ([yshift=\vseg]module.north) {$\bigoplus$};
				
				% output
				\node [] (output) at ([yshift=\vseg]add1.north) {$s_{i+1}$};
				
				\draw[->, thick] (input.north) to node [auto] {} (ln.south);
				\draw[->, thick] (ln.north) to node [auto] {} (module.south);
				\draw[->, thick] (module.north) to node [auto] {} (add1.south);
				\draw[->, thick] (add1.north) to node [auto] {} (output.south);
				%\draw [->, thick ,rounded corners] (input.east) -- ([xshift=2em]input.east)  -- ([xshift=2.1em,yshift=0.1em]add1.east) -- ([yshift=1pt]add1.east);
				\draw [->, thick ,rounded corners] (input.east) -- ([xshift=2em]input.east)  -- ([xshift=2em,yshift=6.55em]input.east) -- ([yshift=1pt]add1.east);

        \node [] (post_input) at (4,0) {$s_i$};
        \node [] (Post) at ([yshift=-1em]post_input.south) {(b) Post-norm};
				
				% module
				\node[layernode,  fill=red!15] (post_module) at ([yshift=\vseg]post_input.north) {\footnotesize{\ \ SubLayer\ \ }};
				
				% residual
        \node[ inner sep=0pt, minimum width=0.15\wnode, minimum height=0.1\hnode] (post_add1) at ([yshift=\vseg]post_module.north) {$\bigoplus$};

        % ln
				\node[layernode, fill=red!15] (post_ln) at ([yshift=\vseg]post_add1.north) {\footnotesize{LayerNorm}};
				
				% output
				\node [] (post_output) at ([yshift=\vseg]post_ln.north) {$s_{i+1}$};
				
				\draw[->, thick] (post_input.north) to node [auto] {} (post_module.south);
				\draw[->, thick] (post_module.north) to node [auto] {} (post_add1.south);
				\draw[->, thick] (post_add1.north) to node [auto] {} (post_ln.south);
				\draw[->, thick] (post_ln.north) to node [auto] {} (post_output.south);
				%\draw[->, thick ,rounded corners] (post_input.east) -- ([xshift=2em]post_input.east)  -- ([xshift=2.1em,yshift=0.1em]post_add1.east) -- ([yshift=1pt]post_add1.east);
				\draw[->, thick ,rounded corners] (post_input.east) -- ([xshift=2em]post_input.east)  -- ([xshift=2em,yshift=4.4em]post_input.east) -- ([yshift=1pt]post_add1.east);

				\end{scope}
				\label{fig:prev}
				\end{tikzpicture}
			}	\\
		\end{tabular}
		
	\end{center}
	
	\begin{center}
		\vspace{-0.5em}
		\caption{Pre-norm and Post-norm sub-layer architectures.}
		\label{fig:prev-norm}
		\vspace{-1.5em}
	\end{center}
\end{figure}

We start with a description of deep Transformer.  In Transformer \cite{vaswani2017attention}, the encoder takes a sequence of words $\{x_{1},...,x_{n}\}$ as input. The input is first transformed into a sequence of embeddings $\{w_{1} + p_{1},...,w_{n} + p_{n}\}$, where $w_{k}$ is a word embedding and $p_{k}$ is a positional embedding. Then, the embedding sequence is fed into a stack of $N$ identical layers. Each layer consists of two stacked sub-layers: a multi-head self-attention sub-layer and a feed-forward sub-layer. The decoder shares a similar architecture as the encoder but possesses an encoder-decoder attention sub-layer to capture the mapping between two languages.

For a deep model, layer normalization networks and layer-wise connections are needed, following the previous work of \citet{bapna-etal-2018-training} and \citet{wang-etal-2019-learning}.

\begin{itemize}

\item \textbf{Pre-Norm Residual Networks}. We make a residual connection \cite{he2016deep} and a layer normalization unit \cite{lei2016layer} at the input of each sub-layer. The output of the sub-layer is defined to be:

\begin{displaymath}
s_{i+1} = s_{i} + \textrm{SubLayer}(\textrm{LayerNorm}(s_i))
\end{displaymath}

where $s_i$ and $s_{i+1}$ are the output of sub-layers $i$ and $i+1$.  See Figure \ref{fig:prev-norm} (a) for the architecture of a pre-norm sub-layer. Pre-norm residual network has been found to be more efficient for back-propagation over a large number of layers than the post-norm architecture \cite{wang-etal-2019-learning,li-etal-2019-niutrans}.

\item \textbf{Dense Connections}. Direct layer connections can make easy access to distant layers in the stack \cite{wang-etal-2018-multi-layer,bapna-etal-2018-training}. Let $\{y_1,...,y_N\}$ be the output of the stacked layers. We define a network $G(y_1,...,y_{j-1})$ that reads all layer output vectors prior to layer $j$ and generates a new vector. Then, $G(y_1,...,y_{j-1})$ is regarded as a part of the input of layer $j$. In this way, we create direct connections from layers $\{1,...,j-1\}$ to layer $j$. For $G(\cdot)$, we choose a linear model as in \cite{wang-etal-2019-learning}.

\end{itemize}

\section{Why do Deep Models Help?}
\label{sec:why}

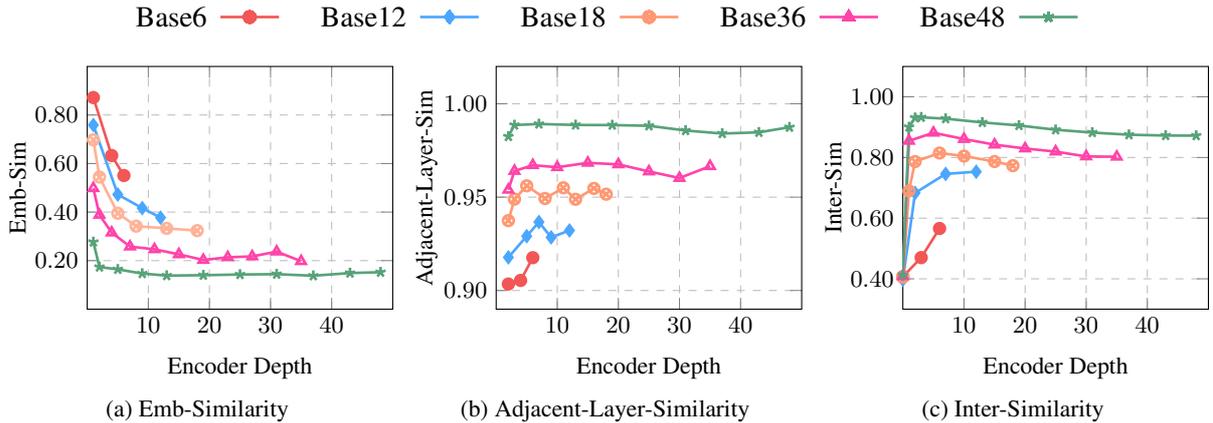
\begin{figure*}[t!]
  \centering
  \begin{tikzpicture}{
    \tikzstyle{layernode} = [draw, thin, rounded corners=1pt, inner sep=3pt,  minimum width=22\wnode, minimum height=0.8\hnode]
    %\node [] (Base) at (0,0) {};
    %\node [] (Base6) at ([xshift=4.5em,yshift=0.7em]Base) {Base6};
    \node [] (Base6) at (0,0) {Base6};
    %\node[layernode] (ln) at ([xshift=15em]Base6) {};
    \node [] (Base12) at ([xshift=5em]Base6.east) {Base12};
    \node [] (Base18) at ([xshift=5em]Base12.east) {Base18};
    \node [] (Base36) at ([xshift=5em]Base18.east) {Base36};
    \node [] (Base48) at ([xshift=5em]Base36.east) {Base48};
    \draw[myred!80,line width=1pt] ([xshift=0.1em]Base6.east) -- plot[mark=otimes*]([xshift=1.1em]Base6.east) -- ([xshift=2.1em]Base6.east);
    \draw[myblue!80,line width=1pt] ([xshift=0.1em]Base12.east) -- plot[mark=diamond*]([xshift=1.1em]Base12.east) -- ([xshift=2.1em]Base12.east);
    \draw[myorange!80,line width=1pt] ([xshift=0.1em]Base18.east) -- plot[mark=otimes]([xshift=1.1em]Base18.east) -- ([xshift=2.1em]Base18.east);
    \draw[mypurple!80,line width=1pt] ([xshift=0.1em]Base36.east) -- plot[mark=triangle]([xshift=1.1em]Base36.east) -- ([xshift=2.1em]Base36.east);
    \draw[mygreen!80,line width=1pt] ([xshift=0.1em]Base48.east) -- plot[mark=star]([xshift=1.1em]Base48.east) -- ([xshift=2.1em]Base48.east);
  }
  \end{tikzpicture}
  \centering
  \subfloat[Emb-Similarity]{
   \centering
   \begin{tikzpicture}
    \footnotesize{
      \begin{axis}[
      ymajorgrids,
  xmajorgrids,
  grid style=dashed,
      width=.35\textwidth,
      height=.30\textwidth,
      legend style={at={(0.42,0.63)}, anchor=south west},
      xlabel={\footnotesize{Encoder Depth}},
      ylabel={\footnotesize{Emb-Sim}},
      ylabel style={yshift=-1em},xlabel style={yshift=0.0em},
      yticklabel style={/pgf/number format/precision=2,/pgf/number format/fixed zerofill},
      ymin=0,ymax=1.0, ytick={0.2, 0.4, 0.6, 0.8},
      xmin=0,xmax=50,xtick={10,20,30,40},
      legend style={yshift=-6pt, legend columns=2,legend plot pos=right,font=\scriptsize,cells={anchor=west}}
      ]

      \addplot[myred!80,mark=otimes*,line width=1pt] coordinates {(1,0.871965587)  (4,0.632949352) (6,0.550674677)};
      %\addlegendentry{\scriptsize Base6}
      \addplot[myblue!80,mark=diamond*,line width=1pt] coordinates {(1,0.758689106) (5,0.472507477)  (9,0.416343808) (12,0.377091814)};
      %\addlegendentry{\scriptsize Base12}
      \addplot[myorange!60,mark=otimes,line width=1pt] coordinates{(1,0.697655082) (2,0.544426024) (5,0.395011961)
       (8,0.341954261) (13,0.333031645)  (18,0.323633075)};
      %\addlegendentry{\scriptsize Base18}
      \addplot[mypurple!80,mark=triangle,line width=1pt] coordinates {(1,0.498284519) (2,0.388617694) (4,0.315523803)
      (7,0.257814914) (11,0.246643215) (15,0.225816488) (19,0.202714607) (23,0.213578179) (27,0.217394873) (31,0.237006098) (35,0.197583184)};
      %\addlegendentry{\scriptsize Base36}
      \addplot[mygreen!80,mark=star,line width=1pt] coordinates {(1,0.276715934) (2,0.173591197) (5,0.164033324)
       (9,0.146789342) (13,0.138284281) (19,0.139886647) (25,0.142813057) (31,0.144191474) (37,0.137800246) (43,0.148809344) (48,0.152258441)};
      %\addlegendentry{\scriptsize Base48}
      \end{axis}
     }
  \end{tikzpicture}\label{fig:cos-from-embedding}
  }
  \subfloat[Adjacent-Layer-Similarity]{
  \centering
  \begin{tikzpicture}
    \footnotesize{
      \begin{axis}[
      ymajorgrids,
  xmajorgrids,
  grid style=dashed,
      width=.35\textwidth,
      height=.30\textwidth,
      legend style={at={(0.42,0.11)}, anchor=south west},
      xlabel={\footnotesize{Encoder Depth}},
      ylabel={\footnotesize{Adjacent-Layer-Sim}},
      ylabel style={yshift=-1em},xlabel style={yshift=0.0em},
      yticklabel style={/pgf/number format/precision=2,/pgf/number format/fixed zerofill},
      ymin=0.89,ymax=1.02, ytick={0.90, 0.95,1},
      xmin=0,xmax=50,xtick={10,20,30,40},
      legend style={yshift=-6pt,legend columns=2, legend plot pos=right,font=\scriptsize,cells={anchor=west}}
      ]

      \addplot[myred!80,mark=otimes*,line width=1pt] coordinates {(2,0.903458607) (4,0.905398012) (6,0.917587481) };
      %\addlegendentry{\scriptsize Base6}
      \addplot[myblue!80,mark=diamond*,line width=1pt] coordinates { (2,0.917758119) (5,0.929079545) (7,0.936668229) (9,0.928464782)  (12,0.932150733)};
      %\addlegendentry{\scriptsize Base12}
      \addplot[myorange!80,mark=otimes,line width=1pt] coordinates{ (2,0.937423558) (3,0.948920987) (5,0.956073604)
       (8,0.949292443)  (11,0.955019674) (13,0.948849878) (16,0.954676659) (18,0.951589427)};
      %\addlegendentry{\scriptsize Base18}
      \addplot[mypurple!80,mark=triangle,line width=1pt] coordinates { (2,0.95394382) (3,0.964007409)  (6,0.96711138)
       (10,0.966033967) (15,0.968376012) (20,0.967553647)  (25,0.963756592) (30,0.960248084)
         (35,0.966572077)};
      %\addlegendentry{\scriptsize Base36}
      \addplot[mygreen!80,mark=star,line width=1pt] coordinates { (2,0.98249208) (3,0.988625755)  (7,0.989191105) (13,0.988671174) (19,0.988605788)
       (25,0.988211682)  (31,0.985691597)  (37,0.984065762) (43,0.984721949) (48,0.98744904)};
      %\addlegendentry{\scriptsize Base48}

      \end{axis}
     }
  \end{tikzpicture}\label{fig:cos-cur-prev}
  }
  \subfloat[Inter-Similarity]{
  \centering
  \begin{tikzpicture}
    \footnotesize{
      \begin{axis}[
      ymajorgrids,
  xmajorgrids,
  grid style=dashed,
      width=.35\textwidth,
      height=.30\textwidth,
      legend style={at={(0.42,0.11)}, anchor=south west},
      xlabel={\footnotesize{Encoder Depth}},
      ylabel={\footnotesize{Inter-Sim}},
      ylabel style={yshift=-1em},xlabel style={yshift=0.0em},
      yticklabel style={/pgf/number format/precision=2,/pgf/number format/fixed zerofill},
      ymin=0.3,ymax=1.1, ytick={0.4, 0.6, 0.8, 1.0},
      xmin=0,xmax=50,xtick={10,20,30,40},
      legend style={yshift=-6pt, legend columns=2,legend plot pos=right,font=\scriptsize,cells={anchor=west}}
      ]

      \addplot[myred!80,mark=otimes*,line width=1pt] coordinates {(0,0.407) (3,0.470) (6,0.566)};
      %\addlegendentry{\scriptsize Base6}
      \addplot[myblue!80,mark=diamond*,line width=1pt] coordinates {(0,0.397243053) (2,0.683173289)  (7,0.745820155)  (12,0.753071179) };
      %\addlegendentry{\scriptsize Base12}
      \addplot[myorange!80,mark=otimes,line width=1pt] coordinates {(0,0.405344933) (1,0.689856999) (2,0.786493652)  (6,0.814697964) (10,0.804633372)  (15,0.786718886)  (18,0.773134542)};
      %\addlegendentry{\scriptsize Base18}
      \addplot[mypurple!80,mark=triangle,line width=1pt] coordinates {(0,0.408574224) (1,0.854825985) (5,0.881507151) (10,0.860796235) (15,0.842295201) (20,0.8296614) (25,0.819406729) (30,0.80345588) (35,0.802439125)};
      %\addlegendentry{\scriptsize Base36}
      \addplot[mygreen!80,mark=star,line width=1pt] coordinates {(0,0.412673861) (1,0.900417936) (2,0.930948329) (3,0.932455492) (7,0.927961659) (13,0.915562642)  (19,0.905633998) (25,0.891421807) (31,0.882193008)  (37,0.875020916)  (43,0.872368582)  (48,0.872029688)};
      %\addlegendentry{\scriptsize Base48}

      \end{axis}
     }
  \end{tikzpicture}\label{fig:sim-individual-global}
  }
  \caption{(a) Similarity of layer $i$ and the input embedding, (b) Similarity of layer $i$ and layer $i-1$, and (c) Inter-Similarity over the validation sequences. Note that 0 represents the embedding layer. }
  \label{fig:cosine}
\end{figure*}

The Transformer encoder (or decoder) is essentially a representation model \cite{vaswani2017attention}. Given a word sequence, a layer generates a distributed representation (i.e., $y_j$) for each position of the sequence. The representation is a mixture of word+position embedding and context embedding. For a simple implementation, only the top-most representation  (i.e., $y_N$) is used for downstream components of the system. Nevertheless, the dense connections can make the lower-level representations directly accessible to the top layers. Hence, the representation model is actually encoded by the set of layer outputs $\{y_1,...,y_N\}$.

For a stronger model, enlarging the size of each layer can fit the objective function with enough capacity. For example, Transformer-Big doubles the layer size of the base model and shows consistent improvements on several MT tasks. But the number of parameters increases quadratically with network width, which poses new difficulties in training such systems and the risk of overfitting. Alternatively, one can stack layers to strengthen the model because more layers offer more representations for the input sentence. Moreover, top-level layers can generate refined representations \cite{iclrGreff2017} by passing the input vectors through more linear and non-linear transformations in different layers.

To study how each layer behaves, we evaluate the change of the representation vector in the stack. To do this, we compute the similarity between the outputs of two layers, as below,

\begin{displaymath}
\textrm{sim}(i,j) = \frac{1}{n} \sum_{l=1}^{n} \textrm{cosine}(y_i(l), y_j(l))
\end{displaymath}

\noindent where $y_i(l)$ (or $y_j(l)$) is the output of layer $i$ (or $j$) for position $l$ of the sequence. $\textrm{sim}(i,j)$ measures the degree of how close the representation vector of layer $i$ is to that of layer $j$.

Figure \ref{fig:cosine}(a) plots $\textrm{sim}(i,0)$ curves for WMT En-De systems of different encoder depths. Here $\textrm{sim}(i,0)$ measures how similar the output of layer $j$ is to the input embedding of the encoder. We see that the similarity keeps going down when we stack more layers. It indicates that the model can learn new representations by using more stacked layers. But the similarity does not converge for shallow models (e.g., 6-layer and 12-layer encoders). It somehow reflects the fact that the shallow models ``want'' more layers to learn something new. More interestingly, deeper models (e.g., encoders of 18 layers or more) make the similarity converge as the depth increases, showing the effect of rendering the need of representation learning fulfilled.

Also, we investigate the similarity between adjacent layers for different systems. Figure \ref{fig:cosine}(b) plots $\textrm{sim}(i,i-1)$ as a function of layer number, which begins with $\textrm{sim}(2,1)$ rather than $\textrm{sim}(1,0)$\footnote{Note that ($\textrm{sim}(1,0)$) of Figure \ref{fig:cosine}(b) are the same with those in Figure \ref{fig:cosine}(a).} for better visualization. The results show that adjacent layers in converged systems have a high similarity. This agrees with previous work on the similarity of attention weights among layers \cite{xiao-etal-2019-sharing} though we study a different issue here. The deeper the models, the higher the similarity between adjacent layers. A natural question is whether we can initialize the higher layers by reusing the parameters of previous layers during training procedure?

Note that the Transformer model is doing something like encoding both contextual information and word information for each position of the sequence. Here, we design the Inter-Sim over the sequence to see how much the representation of a position holds to encode the sequence. For layer $i$, we have

\begin{eqnarray}
\textrm{sim}_{\textrm{in}}(i) & = & \frac{1}{n} \sum_{l=1}^{n} \textrm{cosine}(y_i(l), \bar{y}_i) \nonumber \\
\bar{y}_i & = & \frac{1}{n} \sum_{l=1}^{n} y_i(l) \nonumber
\end{eqnarray}

\noindent where $\bar{y}_i$ is the mean vector of sequence $\{y_i(1),...,y_i(n)\}$ and can be seen as the global representation of the entire sequence. $\textrm{sim}_{\textrm{in}}(i)$ is an indicator of the distance between an individual representation and the global representation. Figure \ref{fig:cosine}(c) shows that the representation of a position tends to be close to the global representation for higher-level layers. This can be seen as smoothing the representations over different positions. A smoothed representation makes the model more robust and is less sensitive to noisy input. This result constitutes evidence that deep models share more global information over different positions of the encoder. Hence, it is easier to access the global representation of the source sequence for decoder and to generate the translation using a global context.

% {\color{red} [problems]} 1. cosine $\to$ norm 2. $\textrm{sim}_{\textrm{in}}(i)$ is not well understood 3. replace $\textrm{sim}_{\textrm{in}}$ with encoder-decoder representation 4. draw figs. 2-4 again!!!  5. misuse of layer and sublayer

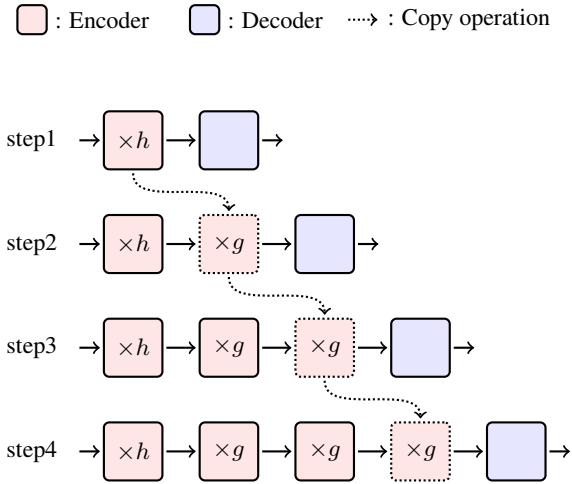
\begin{figure}[t!]
  \centering
  \tikzstyle{encnode} = [rounded corners=2pt,inner sep=4pt,minimum height=2em,minimum width=2em,draw,thick,fill=red!10!white]
  \tikzstyle{copynode} = [rounded corners=2pt,inner sep=4pt,minimum height=2em,minimum width=2em,draw=,densely dotted,thick,fill=red!10!white]
  \tikzstyle{decnode} = [rounded corners=2pt,inner sep=4pt,minimum height=2em,minimum width=2em,draw,thick,fill=blue!10!white]
  \tikzstyle{labelnode} = [rounded corners=2pt,inner sep=4pt,minimum height=1em,minimum width=1em,draw,thick]
  \tikzstyle{cellnode} = [minimum size=0.8em,circle,draw,ublue,thick,fill=white]
  \tikzstyle{standard} = [rounded corners=3pt,thick]
  \centering
    \begin{tikzpicture}
        %step1
        \node [encnode,anchor=west] (enc1) at (0,0) {\footnotesize{$\times h$}};
        \node [decnode,anchor=west] (dec1) at ([xshift=1.2em]enc1.east) {\footnotesize{}};
        \node [anchor=east] (label1) at ([xshift=-1.2em]enc1.west) {\footnotesize{step1}};

        \node [labelnode,anchor=south,fill=red!10!white] (enc) at ([yshift=2.9em]label1.north) {};
        \node [anchor=west] (labelenc) at ([xshift=-0.1em]enc.east) {\footnotesize{: Encoder}};
        \node [labelnode,anchor=west,fill=blue!10!white] (dec) at ([xshift=1em]labelenc.east) {};
        \node [anchor=west] (labeldec) at ([xshift=-0.1em]dec.east) {\footnotesize{: Decoder}};
        \node [anchor=west] (labelcopy) at ([xshift=1.5em]labeldec.east) {\footnotesize{: Copy operation}};
        \draw [->,thick,densely dotted] ([xshift=-1em]labelcopy.west) -- ([xshift=0em]labelcopy.west);

        \draw [->,thick] ([xshift=-0.8em]enc1.west) -- ([xshift=-0.1em]enc1.west);
        \draw [->,thick] ([xshift=0.1em]enc1.east) -- ([xshift=-0.1em]dec1.west);
        \draw [->,thick] ([xshift=0.1em]dec1.east) -- ([xshift=0.8em]dec1.east);

        %step2
        \node [encnode,anchor=north] (enc2) at ([yshift=-1.5em]enc1.south) {\footnotesize{$\times h$}};
        \node [copynode,anchor=west,fill=red!10!white] (enc3) at ([xshift=1.2em]enc2.east) {\footnotesize{$\times g$}};
        \node [decnode,anchor=west] (dec2) at ([xshift=1.2em]enc3.east) {\footnotesize{}};
        \node [anchor=east] (label2) at ([xshift=-1.2em]enc2.west) {\footnotesize{step2}};

        \draw [->,thick] ([xshift=-0.8em]enc2.west) -- ([xshift=-0.1em]enc2.west);
        \draw [->,thick] ([xshift=0.1em]enc2.east) -- ([xshift=-0.1em]enc3.west);
        \draw [->,thick] ([xshift=0.1em]enc3.east) -- ([xshift=-0.1em]dec2.west);
        \draw [->,thick] ([xshift=0.1em]dec2.east) -- ([xshift=0.8em]dec2.east);

        \draw [->,thick,densely dotted] ([yshift=-0.1em]enc1.south) .. controls +(south:0.5) and +(north:0.5) .. ([yshift=0.1em]enc3.north);
        %step3
        \node [encnode,anchor=north] (enc4) at ([yshift=-1.5em]enc2.south) {\footnotesize{$\times h$}};
        \node [encnode,anchor=west] (enc5) at ([xshift=1.2em]enc4.east) {\footnotesize{$\times g$}};
        \node [copynode,anchor=west,fill=red!10!white] (enc6) at ([xshift=1.2em]enc5.east) {\footnotesize{$\times g$}};
        \node [decnode,anchor=west] (dec3) at ([xshift=1.2em]enc6.east) {\footnotesize{}};
        \node [anchor=east] (label3) at ([xshift=-1.2em]enc4.west) {\footnotesize{step3}};

        \draw [->,thick] ([xshift=-0.8em]enc4.west) -- ([xshift=-0.1em]enc4.west);
        \draw [->,thick] ([xshift=0.1em]enc4.east) -- ([xshift=-0.1em]enc5.west);
        \draw [->,thick] ([xshift=0.1em]enc5.east) -- ([xshift=-0.1em]enc6.west);
        \draw [->,thick] ([xshift=0.1em]enc6.east) -- ([xshift=-0.1em]dec3.west);
        \draw [->,thick] ([xshift=0.1em]dec3.east) -- ([xshift=0.8em]dec3.east);

        \draw [->,thick,densely dotted] ([yshift=-0.1em]enc3.south) .. controls +(south:0.5) and +(north:0.5) .. ([yshift=0.1em]enc6.north);

        %step4
        \node [encnode,anchor=north] (enc7) at ([yshift=-1.5em]enc4.south) {\footnotesize{$\times h$}};
        \node [encnode,anchor=west] (enc8) at ([xshift=1.2em]enc7.east) {\footnotesize{$\times g$}};
        \node [encnode,anchor=west] (enc9) at ([xshift=1.2em]enc8.east) {\footnotesize{$\times g$}};
        \node [copynode,anchor=west,fill=red!10!white] (enc10) at ([xshift=1.2em]enc9.east) {\footnotesize{$\times g$}};
        \node [decnode,anchor=west] (dec4) at ([xshift=1.2em]enc10.east) {\footnotesize{}};
        \node [anchor=east] (label3) at ([xshift=-1.2em]enc7.west) {\footnotesize{step4}};

        \draw [->,thick] ([xshift=-0.8em]enc7.west) -- ([xshift=-0.1em]enc7.west);
        \draw [->,thick] ([xshift=0.1em]enc7.east) -- ([xshift=-0.1em]enc8.west);
        \draw [->,thick] ([xshift=0.1em]enc8.east) -- ([xshift=-0.1em]enc9.west);
        \draw [->,thick] ([xshift=0.1em]enc9.east) -- ([xshift=-0.1em]enc10.west);
        \draw [->,thick] ([xshift=0.1em]enc10.east) -- ([xshift=-0.1em]dec4.west);
        \draw [->,thick] ([xshift=0.1em]dec4.east) -- ([xshift=0.8em]dec4.east);

        \draw [->,thick,densely dotted] ([yshift=-0.1em]enc6.south) .. controls +(south:0.5) and +(north:0.5) .. ([yshift=0.1em]enc10.north);

    \end{tikzpicture}

  \caption{Shallow-to-deep training process.}
  \label{fig:shallow2deep}
\end{figure}

\section{Shallow-to-Deep Training}
\label{sec:method}

Although we are able to train deep models with the standard methods \cite{wang-etal-2019-learning}, it is obviously a time consuming task to learn a model with too many layers. For example, training a 48-layer system takes us $3\times$ longer time than the 6-layer baseline. As stated in Section \ref{sec:why}, adjacent layers in a deep network are likely to behave in a similar fashion. This observation inspires us to train upper-level layers by reusing the learned parameters of lower-level layers. We call this method shallow-to-deep training (\textsc{Sdt}) because we start with training a shallow model and then train a deeper model on top of it.

\subsection{The Method}

Assume that we have an initial model $A$ with $h$ layers that have already been trained. Now we need to train a new model $B$ with $h + g$ layers ($h \geq g$). Unlike previous work, we do not train all $h + g$ layers from scratch. Instead,

\begin{itemize}
\item We copy the parameters of the first $h$ layers from $A$ to $B$.
\item We then copy the parameters of the $g$ top-most layers from $A$ to $B$.
\end{itemize}

Model $A$ can be seen as a good starting point of model $B$. After initializing of the model, we continue training model $B$ as usual. The training can converge faster because the model initialization tends to place the parameters in regions of the parameter space that generalize well. See Figure \ref{fig:shallow2deep} for an illustration of the method. In this work, we use the same step to learn from shallow to deep. For example, we train a 6-layer model, and then a 12-layer model, and then a 18-layer model, and so on.

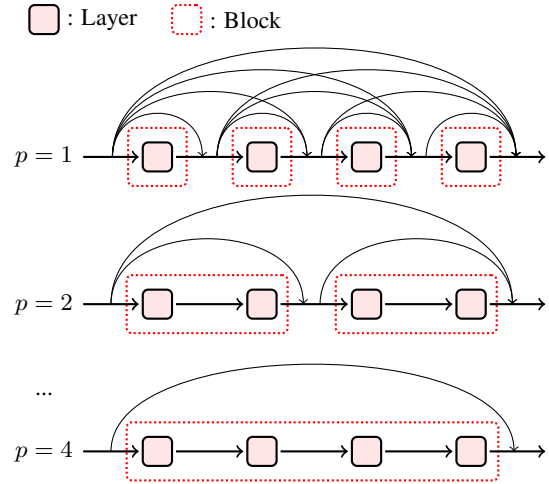
\begin{figure}[t!]
  \centering
  \tikzstyle{encnode} = [rounded corners=2pt,inner sep=3pt,minimum height=1em,minimum width=1em,draw,thick,fill=red!10!white]
  \tikzstyle{blocknode1} = [rounded corners=2pt,inner sep=3pt,minimum height=2em,minimum width=2em,draw,red,thick,densely dotted]
  \tikzstyle{blocknode2} = [rounded corners=2pt,inner sep=3pt,minimum height=2em,minimum width=5.5em,draw,red,thick,densely dotted]
  \tikzstyle{blocknode3} = [rounded corners=2pt,inner sep=3pt,minimum height=2em,minimum width=12.7em,draw,red,thick,densely dotted]
  \tikzstyle{copynode} = [rounded corners=2pt,inner sep=4pt,minimum height=2em,minimum width=2em,draw=,densely dotted,thick,fill=red!10!white]

  \tikzstyle{labelnode} = [rounded corners=2pt,inner sep=4pt,minimum height=1em,minimum width=1em,draw,thick]
  \tikzstyle{cellnode} = [minimum size=0.8em,circle,draw,ublue,thick,fill=white]
  \tikzstyle{standard} = [rounded corners=3pt]
  \centering
    \begin{tikzpicture}
        %p=1
        \node [encnode,anchor=west] (enc1) at (0,0) {\footnotesize{}};
        \node [encnode,anchor=west] (enc2) at ([xshift=2.5em]enc1.east) {\footnotesize{}};
        \node [encnode,anchor=west] (enc3) at ([xshift=2.5em]enc2.east) {\footnotesize{}};
        \node [encnode,anchor=west] (enc4) at ([xshift=2.5em]enc3.east) {\footnotesize{}};
        \node [anchor=east] (label1) at ([xshift=-2em]enc1.west) {\footnotesize{$p=1$}};

        \node [blocknode1,anchor=north] (block1) at ([yshift=0.5em]enc1.north) {\footnotesize{}};
        \node [blocknode1,anchor=north] (block2) at ([yshift=0.5em]enc2.north) {\footnotesize{}};
        \node [blocknode1,anchor=north] (block3) at ([yshift=0.5em]enc3.north) {\footnotesize{}};
        \node [blocknode1,anchor=north] (block4) at ([yshift=0.5em]enc4.north) {\footnotesize{}};

        \node [labelnode,anchor=south,fill=red!10!white] (enc) at ([yshift=3.5em]label1.north) {};
        \node [anchor=west] (labelenc) at ([xshift=-0.1em]enc.east) {\footnotesize{: Layer}};
        \node [labelnode,anchor=west,red,thick,densely dotted] (block) at ([xshift=0.8em]labelenc.east) {};
        \node [anchor=west] (labelblock) at ([xshift=-0.1em]block.east) {\footnotesize{: Block}};

        \draw [->,thick] ([xshift=-2em]enc1.west) -- ([xshift=-0.1em]enc1.west);
        \draw [->,thick] ([xshift=0.1em]enc1.east) -- ([xshift=-0.1em]enc2.west);
        \draw [->,thick] ([xshift=0.1em]enc2.east) -- ([xshift=-0.1em]enc3.west);
        \draw [->,thick] ([xshift=0.1em]enc3.east) -- ([xshift=-0.1em]enc4.west);
        \draw [->,thick] ([xshift=0.1em]enc4.east) -- ([xshift=2em]enc4.east);

        \draw [->,standard] ([xshift=-0.5em]block1.west) .. controls +(90:2em) and +(90:2em) .. ([xshift=0.5em]block1.east);
        \draw [->,standard] ([xshift=-0.5em]block1.west) .. controls +(90:3em) and +(90:3em) .. ([xshift=0.5em]block2.east);
        \draw [->,standard] ([xshift=-0.5em]block1.west) .. controls +(90:4em) and +(90:4em) .. ([xshift=0.5em]block3.east);
        \draw [->,standard] ([xshift=-0.5em]block1.west) .. controls +(90:5em) and +(90:5em) .. ([xshift=0.5em]block4.east);

        \draw [->,standard] ([xshift=-0.5em]block2.west) .. controls +(90:2em) and +(90:2em) .. ([xshift=0.5em]block2.east);
        \draw [->,standard] ([xshift=-0.5em]block2.west) .. controls +(90:3em) and +(90:3em) .. ([xshift=0.5em]block3.east);
        \draw [->,standard] ([xshift=-0.5em]block2.west) .. controls +(90:4em) and +(90:4em) .. ([xshift=0.5em]block4.east);

        \draw [->,standard] ([xshift=-0.5em]block3.west) .. controls +(90:2em) and +(90:2em) .. ([xshift=0.5em]block3.east);
        \draw [->,standard] ([xshift=-0.5em]block3.west) .. controls +(90:3em) and +(90:3em) .. ([xshift=0.5em]block4.east);

        \draw [->,standard] ([xshift=-0.5em]block4.west) .. controls +(90:2em) and +(90:2em) .. ([xshift=0.5em]block4.east);

        %p=2
        \node [encnode,anchor=north] (enc5) at ([yshift=-4em]enc1.south) {\footnotesize{}};
        \node [encnode,anchor=north] (enc6) at ([yshift=-4em]enc2.south) {\footnotesize{}};
        \node [encnode,anchor=north] (enc7) at ([yshift=-4em]enc3.south) {\footnotesize{}};
        \node [encnode,anchor=north] (enc8) at ([yshift=-4em]enc4.south) {\footnotesize{}};
        \node [anchor=east] (label2) at ([xshift=-2em]enc5.west) {\footnotesize{$p=2$}};
        \node [anchor=north] (dot1) at ([yshift=-2em]label2.south) {\footnotesize{...}};

        \node [blocknode2,anchor=north] (block5) at ([xshift=1.7em,yshift=0.5em]enc5.north) {\footnotesize{}};
        \node [blocknode2,anchor=north] (block6) at ([xshift=1.7em,yshift=0.5em]enc7.north) {\footnotesize{}};

        \draw [->,thick] ([xshift=-2em]enc5.west) -- ([xshift=-0.1em]enc5.west);
        \draw [->,thick] ([xshift=0.1em]enc5.east) -- ([xshift=-0.1em]enc6.west);
        \draw [->,thick] ([xshift=0.1em]enc6.east) -- ([xshift=-0.1em]enc7.west);
        \draw [->,thick] ([xshift=0.1em]enc7.east) -- ([xshift=-0.1em]enc8.west);
        \draw [->,thick] ([xshift=0.1em]enc8.east) -- ([xshift=2em]enc8.east);

        \draw [->,standard] ([xshift=-0.5em]block5.west) .. controls +(90:3em) and +(90:3em) .. ([xshift=0.5em]block5.east);
        \draw [->,standard] ([xshift=-0.5em]block5.west) .. controls +(90:5em) and +(90:5em) .. ([xshift=0.5em]block6.east);

        \draw [->,standard] ([xshift=-0.5em]block6.west) .. controls +(90:3em) and +(90:3em) .. ([xshift=0.5em]block6.east);
        %p=4
        \node [encnode,anchor=north] (enca) at ([yshift=-4em]enc5.south) {\footnotesize{}};
        \node [encnode,anchor=north] (encb) at ([yshift=-4em]enc6.south) {\footnotesize{}};
        \node [encnode,anchor=north] (encc) at ([yshift=-4em]enc7.south) {\footnotesize{}};
        \node [encnode,anchor=north] (encd) at ([yshift=-4em]enc8.south) {\footnotesize{}};
        \node [anchor=east] (label2) at ([xshift=-2em]enca.west) {\footnotesize{$p=4$}};

        \node [blocknode3,anchor=north] (block7) at ([xshift=5.3em,yshift=0.5em]enca.north) {\footnotesize{}};

        \draw [->,thick] ([xshift=-2em]enca.west) -- ([xshift=-0.1em]enca.west);
        \draw [->,thick] ([xshift=0.1em]enca.east) -- ([xshift=-0.1em]encb.west);
        \draw [->,thick] ([xshift=0.1em]encb.east) -- ([xshift=-0.1em]encc.west);
        \draw [->,thick] ([xshift=0.1em]encc.east) -- ([xshift=-0.1em]encd.west);
        \draw [->,thick] ([xshift=0.1em]encd.east) -- ([xshift=2em]encd.east);

        \draw [->,standard] ([xshift=-0.5em]block7.west) .. controls +(90:4em) and +(90:4em) .. ([xshift=0.5em]block7.east);

    \end{tikzpicture}

  \caption{Sparse connections between layers.}
  \label{fig:sparse-connection}
\end{figure}

\subsection{Sparse Connections between Layers}

The efficient pass of information plays an important role in training deep models. To this end, one can create direct connections between layers by dense networks. They are found to be necessary to learn strong Transformer systems \cite{bapna-etal-2018-training,wang-etal-2019-learning,wu-etal-2019-depth}. However, an extremely deep model in general results in a large number of such connections and of course a heavy system. For example, a 48-layer system runs $1.87 \times$ slower than the system with no use of dense connections. We cannot even train it using a batch of $2048$ tokens on TITAN V GPUs due to large memory footprint. Instead, we develop a method that resembles the merits of layer-wise connections but is lighter. The idea is pretty simple: we group every $p$ layers to form a layer block and make connections between layer blocks. The connections between blocks are created in the standard way as used in dense networks (see Section \ref{sec:background}). Here $p$ is a parameter to control the connection density. For example, $p=1$ means dense networks, and $p=\infty$ means networks with no layer-wise connections \footnote{Residual connections are used by default in this work.}. See Figure \ref{fig:sparse-connection} for example networks with block/layer-wise connections.
% In this work, we set the value of $p$ equal to $h$ and $g$ for robust training. The reason is easy to comprehensive that NMT is a model based on encoder-decoder paradigm, unlike the ResNets and the BERT. During training a deep model from shallow-to-deep, the decoder need to gradually finetune the weights in each stage.

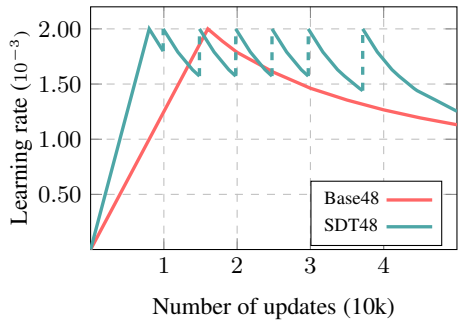
\begin{figure}
  \centering
  \begin{tikzpicture}
    \footnotesize{
      \begin{axis}[
      ymajorgrids,
	xmajorgrids,
	grid style=dashed,
      width=.40\textwidth,
      height=.30\textwidth,
      legend style={at={(0.60,0.08)}, anchor=south west},
      xlabel={\footnotesize{Number of updates (10k)}},
      ylabel={\footnotesize{Learning rate  (\scriptsize{$10^{-3}$)}}},
      ylabel style={yshift=-1em},xlabel style={yshift=0.0em},
      yticklabel style={/pgf/number format/precision=2,/pgf/number format/fixed zerofill},
      ymin=0,ymax=2.2, ytick={0.5, 1, 1.5, 2},
      xmin=0,xmax=5,xtick={1,2,3,4},
      legend style={yshift=-6pt, legend plot pos=right,font=\scriptsize,cells={anchor=west}}
      ]

      \addplot[red!60,line width=1.25pt] coordinates {(0,0) (1.6,2) (1.8,1.888) (2,1.787) (2.5,1.606) (3,1.462) (3.5,1.3549) (4,1.266) (4.5,1.193) (5,1.131)};
      \addlegendentry{\scriptsize Base48}
      %\addplot[red,line width=1.25pt] coordinates {(0,0) (8000,0.002) (10000,0.00179) (12000,0.00163) (12950,0.001572)};
      \addplot[teal!70,line width=1.25pt] coordinates {(0,0) (0.8,2) (0.9906,1.7983)};
      %\addplot[red,line width=1.25pt] coordinates {(0,0) (8000,0.002) (9906,0.0017983)};
      \addplot[teal!70,dashed,line width=1.25pt] coordinates {(0.9906,1.7983) (0.9906,2)};
      \addplot[teal!70,line width=1.25pt] coordinates {(0.9906,2) (1.1906,1.79) (1.3906,1.63) (1.4856,1.572)};
      \addplot[teal!70,dashed,line width=1.25pt] coordinates {(1.4856,1.572) (1.4856,2)};
      \addplot[teal!70,line width=1.25pt] coordinates {(1.4856,2) (1.6856,1.79) (1.8856,1.63) (1.9806,1.572)};
      \addplot[teal!70,dashed,line width=1.25pt] coordinates {(1.9806,1.572) (1.9806,2)};
      \addplot[teal!70,line width=1.25pt] coordinates {(1.9806,2) (2.1806,1.79) (2.3806,1.63) (2.4756,1.572)};
      \addplot[teal!70,dashed,line width=1.25pt] coordinates {(2.4756,1.572) (2.4756,2)};
      \addplot[teal!70,line width=1.25pt] coordinates {(2.4756,2) (2.6756,1.79) (2.8756,1.63) (2.9706,1.572)};
      \addplot[teal!70,dashed,line width=1.25pt] coordinates {(2.9706,1.572) (2.9706,2)};
      \addplot[teal!70,line width=1.25pt] coordinates {(2.9706,2) (3.1706,1.79) (3.3706,1.63) (3.4656,1.572) (3.6706,1.4602) (3.7136,1.44)};
      \addplot[teal!70,dashed,line width=1.25pt] coordinates {(3.7136,1.44) (3.7136,2)};
      \addplot[teal!70,line width=1.25pt] coordinates {(3.7136,2) (3.9136,1.79) (4.1136,1.63) (4.2086,1.572) (4.4136,1.4602) (4.4566,1.44) (4.7000,1.3574) (5.0000,1.2531)};
      \addlegendentry{\scriptsize SDT48}

      \end{axis}
     }
  \end{tikzpicture}
  \caption{The learning rate schedule of each stacking.}\label{fig:lr}
\end{figure}

\subsection{Learning Rate Restart}

The design of the learning rate schema is one of the keys to the success of Transformer. For example, \citet{vaswani2017attention} designed a method to warm up training for a number of training steps and then decrease the learning rate. In this work, shallow-to-deep training breaks the training process because we need to switch to a deeper model with initialization at some training steps. We found in our experiments that deep models could not be trained efficiently in the standard way because we had a small learning rate for a newly stacked model in late training steps.

We develop a new method to ensure that the model can be trained using a proper learning rate at every point of switching to a deeper model. We divide the training into a number of stages. Each of them is associated with a deeper model.

\begin{itemize}
  \item For the first $\omega$ stages, the model is trained with a linear-warmup learning rate ($lr$) as described in \cite{vaswani2017attention}.
  \begin{equation}
    lr = d^{-0.5}_{\textrm{model}} \cdot step\_num \cdot warmup\_steps^{-1.5} \nonumber
  \end{equation}
  \item For each of the following stages, the learning rate of a newly stacked model declines from the max-point with an inverse squared root of the current step.
  \begin{equation}
    lr = d^{-0.5}_{\textrm{model}} \cdot step\_num^{-0.5} \nonumber
  \end{equation}
  At the beginning of the stage, we reset the number of training steps.
\end{itemize}

Here $step\_num$ and $warmup\_steps$ are the current training step number and the warmup-step number. $d_{\textrm{model}}$ is the size of the layer output. See Figure \ref{fig:lr} for a comparison of different learning schemas.

\section{Experiments}
\label{sec:exp}

We report the experimental results on two widely used benchmarks - WMT'16 English-German (En-De) and WMT'14 English-French (En-Fr).
\subsection{Data}

For the En-De task, we used the same preprocessed data with \cite{vaswani2017attention,ott-etal-2019-fairseq,wang-etal-2019-learning}, consisting of approximate $4.5$M tokenized sentence pairs. All sentences were segmented into sequences of sub-word units \cite{sennrich-subword-neural} with $32$K merge operations using a vocabulary shared by source and target sides. We selected \textit{newstest2012}+\textit{newstest2013} as validation data and \textit{newstest2014} as test data.

For the En-Fr task, we replicated the setup of \citet{vaswani2017attention} with 36M training sentence pairs from WMT14. We validated the En-Fr system on the union set of \textit{newstest2012} and \textit{newstest2013}, and tested it on \textit{newstest2014}. We filtered out sentences of more than 200 words and generated a shared vocabulary with $40$K merge operations on both source and target side.

We re-merged sub-word units to form complete words in the final output. For comparable results with previous work \cite{wu2016google,gehring2017convs2s,vaswani2017attention}, we also adopted compound split for En$\rightarrow$De. We reported case-sensitive tokenized BLEU scores for both En-De and En-Fr tasks, and sacrebleu\footnote{BLEU+case.mixed+numrefs.1+smooth.exp+tok.13a\\+version.1.2.12} scores for both En-De and En-Fr tasks. The results were the mean of three times run with different random seeds.

\begin{table*}[!htbp]
  \centering
  \setlength{\tabcolsep}{2.7pt}
  \small
  \begin{tabular}{l rrccc rrccc}
  \toprule
  \multirow{2}{*}{\textbf{Systems}} & \multicolumn{5}{c }{\textbf{WMT En-De}}  & \multicolumn{5}{c}{\textbf{WMT En-Fr}} \\
  \cmidrule(r){2-6} \cmidrule(r){7-11} 
  &{\bf Params}&{\bf Time}&{\bf Speedup}&{\bf BLEU}&{\bf Sacrebleu} &{\bf Params}&{\bf Time}&{\bf Speedup}&{\bf BLEU}&{\bf Sacrebleu}\\

  \midrule
  \citet{vaswani2017attention} (Big)    &213M   &N/A    &N/A        &28.40     &N/A      &222M    &N/A     &N/A         &41.00      &N/A  \\
  \citet{shaw-etal-2018-self} (Big)     &210M   &N/A    &N/A        &29.20     &N/A      &222M    &N/A     &N/A         &41.30      &N/A  \\
  \citet{ott-EtAl:2018:WMT} (Big)       &210M   &N/A    &N/A        &29.30     &28.6     &222M    &N/A     &N/A         &43.20      &41.4 \\
  \citet{wu-etal-2019-depth} (Big)      &270M   &N/A    &N/A        &29.92     &N/A      &281M    &N/A     &N/A         &43.27      &N/A  \\
  \citet{wang-etal-2019-learning} (Deep)&137M   &N/A    &N/A        &29.30     &N/A      &N/A     &N/A     &N/A         &N/A        &N/A  \\
  \citet{wei2004multiscale} (Deep)      &272M   &N/A    &N/A        &30.19     &N/A      &N/A     &N/A     &N/A         &N/A        &N/A  \\
  \citet{wei2004multiscale} (Big+Deep)       &512M   &N/A    &N/A        &30.56     &N/A      &N/A     &N/A     &N/A         &N/A        &N/A  \\
  \cdashline{1-11}     
  Base (Pre-Norm)                       &63M   &4.79   &N/A         &27.05     &26.0     &67M   &27.11     &N/A         &41.00      &39.2 \\
  Big (Pre-Norm)                        &210M  &36.05  &N/A         &28.79     &27.7     &222M  &97.51     &N/A         &42.40      &40.6 \\
  \midrule         
  Deep-24L                              &118M  &8.66   &0           &28.95     &27.8     &124M   &48.43    &0           &42.40      &40.6 \\
  \textsc{Sdt}-24L                      &118M  &6.16   &28.92\%     &29.02     &27.9     &124M   &33.81    &30.10\%     &42.42      &40.6 \\
  Deep-RPR-24L                          &118M  &9.80   &0           &29.39     &28.3     &124M   &55.32    &0           &42.67      &40.9 \\
  \textsc{Sdt}-RPR-24L                  &118M  &6.71   &31.53\%     &29.39     &28.3     &124M   &37.59    &32.05\%     &42.69      &40.9 \\
  \cdashline{1-11}           
  Deep-48L                              &194M  &16.38  &0           &29.44     &28.3     &199M   &90.85    &0           &42.75      &41.0 \\
  \textsc{Sdt}-48L                      &194M  &10.65  &35.02\%     &29.60     &28.5     &199M   &55.35    &39.08\%     &42.82      &41.0 \\
  Deep-RPR-48L                          &194M  &19.58  &0           &30.03     &28.8     &199M   &116.92   &0           &43.08      &41.3 \\
  \textsc{Sdt}-RPR-48L                  &194M  &11.75  &\bf 39.98\% &\bf 30.21 &\bf 29.0 &199M   &64.46    &\bf 44.90\% &\bf 43.29  &\bf 41.5 \\
  \cdashline{1-11}
  Deep-24L (Big)                    &437M  &37.41  &0           &29.90     &28.7     &N/A    &N/A      &N/A         &N/A        &N/A \\
  \textsc{Sdt}-24L (Big)            &437M  &18.31  &47.41\%     &29.93     &28.7     &N/A    &N/A      &N/A         &N/A        &N/A \\  
  Deep-RPR-24L (Big)                    &437M  &38.80  &0           &30.40     &29.2     &N/A    &N/A      &N/A         &N/A        &N/A \\
  \textsc{Sdt}-RPR-24L (Big)            &437M  &18.51  &\bf 52.30\% &\bf 30.46 &\bf 29.3 &N/A    &N/A      &N/A         &N/A        &N/A \\
  \bottomrule 
  \end{tabular}
  \caption{Results of deep models on WMT14 En-De and WMT14 En-Fr tasks by the model parameters [million], training costs [hours], acceleration rates [$\%$], BLEU scores [$\%$], $\bigtriangleup$ BLEU [$\%$] and Sacrebleu scores [$\%$].}\label{tab:main-results}
\end{table*}

\subsection{Model Settings}

Our implementation was based on Fairseq \cite{ott-etal-2019-fairseq}. For training, we used Adam optimizer \cite{kingma2014adam} with $\beta_1=0.9$, $\beta_2=0.997$, and $\epsilon=10^{-8}$. We adopted the same learning rate schedule as the latest implementation of Tensor2Tensor\footnote{\href{https://github.com/tensorflow/tensor2tensor}{https://github.com/tensorflow/tensor2tensor}}. For deep models, the learning rate ($lr$) first increased linearly for $warmup=8,000$ steps from $1e^{-7}$ to
$2e^{-3}$. After warmup, the learning rate decayed proportionally to the inverse square root of the current step. For our \textsc{Sdt} method presented in Section \ref{sec:method}, we set $h=g=p=6$ on both the WMT En-De and En-Fr tasks. For a stronger system, we employed relative position representation (RPR) to strengthen the position embedding model \cite{shaw-etal-2018-self}. We only used the relative key in each layer.

We batched sentence pairs by approximate length, and limited input/output tokens per batch to $4,096$/GPU. Following the method of \cite{wang-etal-2019-learning}, we accumulated every two steps for a better batching. This resulted in approximately $56,000$ source and $56,000$ target tokens per training batch. The deep models were updated for 50k steps on the En-De task and 150k steps on the En-Fr task.  All models were trained on 8 NVIDIA TITAN V GPUs with mix-precision accelerating. For fair comparison, we trained the deep Pre-Norm Transformer with the same settings reported in \citet{wang-etal-2019-learning}. And all results are the average of three times running with different random seeds. We chose different hyper-parameter settings for the models. 

\begin{itemize}

  \item Base/Deep Model. The hidden layer size of self-attention was $512$, and the size of feed forward inner-layer was $2,048$. Also, we used 8 heads for attention. For training, we set all dropout to $0.1$, including residual dropout, attention dropout, relu dropout. Label smoothing $\epsilon_{ls}=0.1$ was applied to enhance the generation ability of the model.
  
  \item Big Model. We used the same architecture as Transformer-Base but with a larger hidden layer size $1,024$, more attention heads ($16$), and a larger feed forward inner-layer ($4,096$ dimensions). The residual dropout was set to $0.3$ for the En-De task and $0.1$ for the En-Fr task. Additionally, the same depth (6) on both encoder and decoder side with Base model. For deeper Big model, we only change the encoder depth.
  
  \end{itemize}

For evaluation, we averaged the last 5 consecutive checkpoints which were saved per training epoch on all WMT models. For all datasets, the length penalty was set to 0.6 and the beam size was set to 4.

\subsection{Results}

Table \ref{tab:main-results} summarizes the training cost and the translation quality on the WMT En-De and En-Fr tasks. First, we compare deep Transformer systems (Pre-Norm) with previously reported systems. Deep Transformer brings substantial improvements than big counterparts within the same experiment settings\footnote{The models of \citet{ott-EtAl:2018:WMT} were trained on 128 GPUS. And \citet{wu-etal-2019-depth} trained their networks for $800,000$ steps.}. In addition, our \textsc{Sdt} method enables efficient training for deeper networks with no loss in BLEU. As we can see from Table \ref{tab:main-results}, the systems trained with the \textsc{Sdt} method achieve comparable or even higher BLEU scores with their baselines, and the training costs are much less.

Another finding is that systems encoding relative position representation in the same encoder depth outperform their baselines by $0.44-0.59$ BLEU points on the En-De task, indicating that relative position representation can further strengthen deep Transformer. Similarly, it is observed that \textsc{Sdt} can also speed up the RPR enhanced systems with no loss of translation quality. The speedup is larger for deeper models that the \textsc{Sdt} method speeds up the training of 48-layer systems by $35.02\%-39.98\%$. Surprisingly, both \textsc{Sdt}-48L and \textsc{Sdt}-RPR-48L achieve modest BLEU improvements compared with the baseline. This indicates that the benefit from enlarging encoder depth gradually decreases and our method alleviates the overfitting problem when the model is extremely deep.

\begin{table}[t!]
  \small
  \begin{center}
  \begin{tabular}{ccc}
  \toprule
  \textbf{Reset-lr} & \textbf{Copy-Initialization} & \textbf{BLEU} \\
  \midrule
  % \multirow{3}*{SDT-RPR-48L}
  $\times$      & $\times$      & 27.33 \\
  \checkmark    & $\times$      & 27.98 \\
  $\times$      & \checkmark    & 29.93 \\
  \checkmark    & \checkmark    & \textbf{30.21} \\

  \bottomrule
  \end{tabular}
  \end{center}
  \caption{Effect of learning rate schema and copy-initialization strategy.}\label{tab:ablation}
\end{table}

\begin{table}
  \small
  \begin{center}
  \begin{tabular}{lcc}
  \toprule
  \textbf{Copy-Initialization} & \textbf{Interval} & \textbf{BLEU} \\
  \midrule
  % \multirow{3}*{SDT-RPR-48L}
  Top only          &$g=6$       & 29.20 \\
  Interpolation     &$g=6$       & 29.46 \\
  $g$ Top-most      &$g=6$       & \textbf{30.21} \\

  \bottomrule
  \end{tabular}
  \end{center}
  \caption{The comparison of our method with other copy initialization strategies.}\label{tab:copy_initialization}
\end{table}

To further validate the effectiveness of \textsc{Sdt} method, we experimented on 24-layer Big models. Through Table \ref{tab:main-results} we see that, it achieves up to $52.30\%$ speedup and match the performance with learning from scratch. Note that the training time of Deep-24L (Big) and Deep-RPR-24L (Big) are 37.41 and 38.80 hours respectively because the models were only optimized by 50k steps and they converged on the validation set. The finding here is similar with \citet{wang-etal-2019-learning}'s work that deep encoders can speed up the training process with a large learning rate. In addition, the BLEU score of \textsc{Sdt}-RPR-24L (Big) is $30.46$, which matches with the state-of-the-art within less parameters and training cost. Another finding here is the speedup of \textsc{Sdt} training may gets larger when the model architecture gets more complex.

The similar phenomenon is observed in the WMT En-Fr task, a much larger dataset than that of the En-De task. The results in Table \ref{tab:main-results} show the effectiveness of our \textsc{Sdt} method. It accelerates the training procedure by $44.90\%$ with nearly $0.2$ BLEU improvement on a 48-layer RPR system. The larger the dataset is, the greater the speedup will be. Note that our 48-layer deep system matches the state-of-the-art reported in \cite{ott-EtAl:2018:WMT,wu2019pay,wu-etal-2019-depth} within 8 GPUs training. We will furthermore verify whether there is an another improvement when we switch to a much larger batching schema. However, due to the large size of En-Fr dataset, optimizing a 24-layer Big model is quite time consuming, thus we have not finished the training yet. The experimental results can be found in our codebase soon. In addition, another benifit brought by \textsc{Sdt} is that we can efficiently build the ensemble system given an already optimized model, we will show more details in our codebase. 

% Deep-RPR-48L fails to train due to gradient vanishing/exploding. As expected, the Sparse-RPR-48L system works well and achieves a BLEU score of 43.08 points. Our SDT-RPR-48L model outperforms Sparse-RPR-48L by 0.1 BLEU scores. This agrees with the result reported in \cite{wu2019pay}. The training speed up is $1.44 \times$.

% As expected, the sparse connections between layers are efficient, and the increased model size is negligible. 
% Note that the training process may crash due to the big variance of gradients norm if we progressively train a system deeper than 30 layers using \textsc{Sdt}. Therefore, we also adopt the sparse connections across each block (see Figure \ref{fig:sparse-connection}) to ease the information transmitting.

\begin{table}[t!]
  \small
  \begin{center}
  \begin{tabular}{lcrr}
  \toprule

  \textbf{Strategy} & \textbf{Interv.} &\textbf{Speedup}& \textbf{BLEU} \\
  \midrule
  % \hline
  $g=3$         &1    &36.7\%               & 29.38                   \\
  $g=6$         &2    &39.9\%               & \textbf{30.21}          \\
  $g=9$         &4    &42.1\%               & 29.47                   \\
  $g=6,9,12,15$ &4    &\textbf{51.8}\%      & 29.78                   \\
  % \hline
  \bottomrule
  \end{tabular}
  \end{center}
  \caption{BLEU scores [$\%$] vs. speedup of different stacking strategies during training.}\label{tab:stack}
\end{table}
\begin{table}[t!]
  \small
  \setlength{\tabcolsep}{4.5pt}
  \begin{center}
  \begin{tabular}{lrrc}
  \toprule

  \textbf{System} &\textbf{Speedup}& \textbf{BLEU} & \textbf{Sacrebleu} \\
  \midrule
  % \hline
  Deep-RPR-30L                 &N/A                & 29.52      & 28.4     \\
  DLCL-RPR-30L                 &ref                & 30.01      & 28.9     \\
  Sparse-RPR-30L               &8.9\%              & 29.90      & 28.8     \\
  \textsc{Sdt}-RPR-30L         &\textbf{42.1}\%    & 29.95      & 28.9     \\
  \textsc{Sdt}-RPR-54L         &N/A                & \bf 30.33  & \bf 29.2 \\
  \textsc{Sdt}-RPR-60L         &N/A                & 30.28      & 29.1     \\
  % \hline
  \bottomrule
  \end{tabular}
  \end{center}
  \caption{Comparison of DLCL and our work.}\label{tab:comparison}
\end{table}

\section{Analysis}
\label{sec:analysis}
\subsection{Ablation Study}

In Table \ref{tab:ablation}, we summarize the effect of resetting the learning rate and copy-initialization strategy in our \textsc{Sdt} method.  We choose \textsc{Sdt}-RPR-48L as our baseline due to its strong performance. We see, first of all, that the copy-initialization plays an important role in our shallow-to-deep training process. For example, the BLEU score decreases dramatically if we stack the model from 6 layers to 12 layers and initialize the new block by a uniform distribution. This can be explained by the fact that every top-most layers are not sufficiently converged in stacking. Another observation is that the reset learning rate can also facilitate the training, enabling the model to learn fast and bringing nearly a +0.3 BLEU improvement. Moreover, we compare different copy-initialization strategies, including initializing the new group by copying the $g$ top-most layers, copying the top-most layer for $g$ times and inserting the layer right after each existing layer in the group. From Table \ref{tab:copy_initialization}, we observe that our copying $g$ top-most strategy achieves best performance.

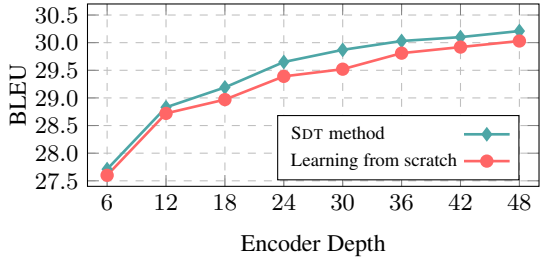
\begin{figure}[!t]
  \centering
  \begin{tikzpicture}
    \footnotesize{
      \begin{axis}[
      ymajorgrids,
  xmajorgrids,
  grid style=dashed,
      width=.47\textwidth,
      height=.25\textwidth,
      legend style={at={(0.42,0.12)}, anchor=south west},
      xlabel={\footnotesize{Encoder Depth}},
      ylabel={\footnotesize{BLEU}},
      ylabel style={yshift=-1em},xlabel style={yshift=0.0em},
      yticklabel style={/pgf/number format/precision=1,/pgf/number format/fixed zerofill},
      ymin=27.4,ymax=30.7, ytick={27.50,28.00,28.50,29.00,29.50,30.00,30.50},
      xmin=4,xmax=50,xtick={6,12,18,24,30,36,42,48},
      legend style={yshift=-6pt, legend plot pos=right,font=\scriptsize,cells={anchor=west}}
      ]

      \addplot[teal!70,mark=diamond*,line width=1pt] coordinates {(6,27.71) (12,28.83) (18,29.19) (24,29.65) (30,29.87) (36,30.03) (42, 30.10) (48, 30.21)};
      \addlegendentry{\scriptsize \textsc{Sdt} method}
      \addplot[red!60,mark=otimes*,line width=1pt] coordinates {(6,27.60) (12,28.72) (18,28.97) (24,29.39) (30,29.52) (36,29.81) (42,29.92) (48, 30.03)};
      \addlegendentry{\scriptsize Learning from scratch}
      \end{axis}
     }
  \end{tikzpicture}
  \caption{The comparison between learning from scratch and \textsc{Sdt} against different encoder depth.}\label{fig:sdt-detph}
\end{figure}

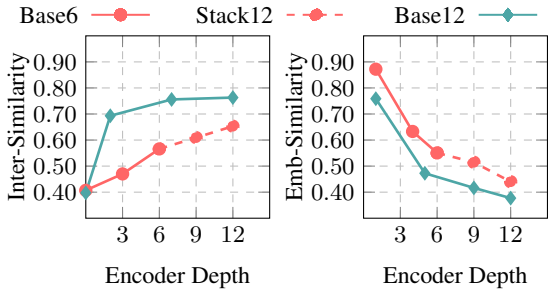
\begin{figure}[!t]
	\centering
	\begin{tikzpicture}{
	  \tikzstyle{layernode} = [draw, thin, rounded corners=1pt, inner sep=3pt,  minimum width=13.2\wnode, minimum height=0.8\hnode]
	  %\node [] (Base) at (0,0) {};
      %\node [] (Base6) at ([xshift=0em,yshift=0.4em]Base) {\small{Base6}};
      \node [] (Base6) at (0,0) {\small{Base6}};
	  \node [] (Stack12) at ([xshift=5em]Base6.east) {\small{Stack12}};
	  \node [] (Base12) at ([xshift=5em]Stack12.east) {\small{Base12}};
	  %\node[layernode] (ln) at ([xshift=8em]Base6) {};
	  \draw[red!60,line width=0.75pt] ([xshift=0.1em]Base6.east) -- plot[mark=otimes*]([xshift=1.1em]Base6.east) -- ([xshift=2.1em]Base6.east);
	  \draw[red!60,dashed,line width=0.75pt] ([xshift=0.1em]Stack12.east) -- plot[mark=otimes*]([xshift=1.1em]Stack12.east) -- ([xshift=2.1em]Stack12.east);
	  \draw[teal!70,line width=0.75pt] ([xshift=0.1em]Base12.east) -- plot[mark=diamond*]([xshift=1.1em]Base12.east) -- ([xshift=2.1em]Base12.east);

	}
	\end{tikzpicture}
	\\
	%\subfloat[Inter-Similarity]{
	 \centering
	 \begin{tikzpicture}{baseline}
	 \footnotesize{
	  \begin{axis}[
	  ymajorgrids,
	  xmajorgrids,
	  grid style=dashed,
	  width=.25\textwidth,
	  height=.25\textwidth,
	  legend style={at={(0.32,0.08)}, anchor=south west},
	  xlabel={\footnotesize{Encoder Depth}},
	  ylabel={\footnotesize{Inter-Similarity}},
	  % xmajorgrids=true,
	  % ymajorgrids=true,
	  % grid style = dashed,
	  ylabel style={yshift=-1.1em},xlabel style={yshift=0.0em},
	  yticklabel style={/pgf/number format/precision=2,/pgf/number format/fixed zerofill},
	  ymin=0.3,ymax=1.0, ytick={0.4, 0.5, 0.6,0.7,0.8,0.9},
	  xmin=0,xmax=15,xtick={3,6,9,12},
	  legend style={yshift=-6pt, legend plot pos=right,font=\scriptsize,cells={anchor=west}}
	  ]

	  \addplot[red!60,mark=otimes*,line width=1pt] coordinates {(0,0.407) (3,0.470) (6,0.566)};
	  \addplot[red!60,mark=otimes*,dashed,line width=1pt] coordinates {(6,0.566) (9,0.60928) (12,0.65306)};
	  \addplot[teal!70,mark=diamond*,line width=1pt] coordinates {(0,0.397243053) (2,0.693173289)  (7,0.755820155)  (12,0.763071179) };

	  \end{axis}
	 }
	 \end{tikzpicture}
	%}
	%\subfloat[Emb-Similarity]{
	\centering
	 \begin{tikzpicture}{baseline}
	 \footnotesize{
	  \begin{axis}[
	  ymajorgrids,
	  xmajorgrids,
	  grid style=dashed,
	  width=.25\textwidth,
	  height=.25\textwidth,
	  legend style={at={(0.32,0.08)}, anchor=south west},
	  xlabel={\footnotesize{Encoder Depth}},
	  ylabel={\footnotesize{Emb-Similarity}},
	  ylabel style={yshift=-1.1em},xlabel style={yshift=0.0em},
	  yticklabel style={/pgf/number format/precision=2,/pgf/number format/fixed zerofill},
	  ymin=0.3,ymax=1.0, ytick={0.4, 0.5, 0.6,0.7,0.8,0.9},
	  xmin=0,xmax=15,xtick={3,6,9,12},
	  legend style={yshift=-6pt, xshift=-2pt,legend plot pos=right,font=\scriptsize,cells={anchor=west}}
	  ]

	  \addplot[red!60,mark=otimes*,line width=1pt] coordinates {(1,0.871965587)  (4,0.632949352) (6,0.550674677)};
	  \addplot[red!60,mark=otimes*, dashed,line width=1pt] coordinates {(6,0.550674677) (9,0.5146) (12,0.44)};
	  \addplot[teal!70,mark=diamond*,line width=1pt] coordinates {(1,0.758689106) (5,0.472507477)  (9,0.416343808) (12,0.377091814)};

	  \end{axis}
	 }
	 \end{tikzpicture}
	%}
	\caption{Inter and Emb Similarity of Stack-12L.}
  \label{fig:sim-stack}
\end{figure}

Also, we investigate the impact of different stacking strategies on translation quality and speedup. Table \ref{tab:stack} shows results of the models trained with different settings of $g$ and training intervals\footnote{Training interval means the training epoch of each newly stacking model}. Row 4 denotes the case that we stack the shallow model in an incremental way. We find that the stacking strategy and its training interval make great impacts on both translation quality and speedup. We need a trade-off to select the ``best'' system in different situations. For example, our default strategy (line 2) obtains the best performance, and the incremental stacking achieves the biggest speedup.

\subsection{Comparison with Previous Work}

Next, we compare the system Transformer-DLCL \cite{wang-etal-2019-learning} with our \textsc{Sdt} system. Table \ref{tab:comparison} shows the BLEU scores of the models trained with DLCL, sparse connection and \textsc{Sdt} based on RPR, respectively. We see that Sparse-RPR-30L can achieve comparable performance with DLCL-RPR-30L using much fewer connections across encoder layers. More interestingly, our \textsc{Sdt}-RPR-30L has a comparable BLEU score with DLCL-RPR-30L, but is $42.1\%$ faster. In addition, we find that \textsc{Sdt}-RPR-54L outperforms \textsc{Sdt}-RPR-30L by 0.38 BLEU scores, but much deeper models cannot gain more benefits. This result indicates that deeper representation models might suffer from the overfitting problem.

Another benefit brought by \textsc{Sdt} method is that we can train a deep Transformer model from a pre-trained model instead of training from scratch. For example, we can begin training from a pre-trained 24-layer system to progressively obtain a 48-layer system. The experimental results in Figure \ref{fig:sdt-detph} verify our conjecture. Except the advantage of accelerating the training, the models trained through the \textsc{Sdt} method can even slightly outperform those training from scratch at almost all encoder depths. This enables us to quickly obtain single systems in different depth from an already trained system, which is efficient to build ensemble systems, especially when the training data is extremely large. 

\begin{figure}[!t]
  \centering
  \vspace{2pt}
  \begin{tikzpicture}
    \footnotesize{
    \begin{axis}[
    ymajorgrids,
    xmajorgrids,
    grid style=dashed,
    width=.47\textwidth,
    height=.25\textwidth,
    legend style={at={(0.43,0.85)}, anchor=south west},
    xlabel={\footnotesize{Encoder Depth}},
    ylabel={\footnotesize{Emb-Similarity}},
    ylabel style={yshift=-1em},xlabel style={yshift=0.0em},
    yticklabel style={/pgf/number format/precision=2,/pgf/number format/fixed zerofill},
    ymin=0,ymax=1.0, ytick={0.2, 0.4, 0.6, 0.8},
    xmin=0,xmax=50,xtick={10,20,30,40},
    legend style={yshift=-6pt, legend columns=2,legend plot pos=right,font=\scriptsize,cells={anchor=west},at={(0.4,0.85)}}
    ]

    \addplot[teal!70,mark=diamond*,line width=1pt] coordinates {(1,0.781430137) (3,0.63762368) (6,0.484653008) (7,0.522484851)  (12,0.375641665) (13,0.441803795) (18,0.357287538) (19,0.431271297)  (24,0.358096015) (25,0.433635634)  (30,0.345990442) (31,0.400078696)  (36,0.302658421) (37,0.407904637)  (42,0.293056381) (43,0.425834608) (48,0.225051624)};
    \addlegendentry{\scriptsize SDT48}
    %\addplot[orange,mark=otimes*,line width=1.25pt] coordinates {(1,0.831430137) (3,0.68762368) (6,0.534653008) (7,0.572484851)  (10,0.485666573) (12,0.415641665) (13,0.491803795) (16,0.420627564) (18,0.407287538) (19,0.481271297) (22,0.434586257)  (24,0.408096015) (25,0.483635634) (27,0.451462954) (30,0.415990442) (31,0.450078696) (33,0.418118834) (36,0.352658421) (37,0.457904637)  (40,0.406968564) (42,0.343056381) (43,0.475834608) (46,0.381614685) (48,0.275051624)};
    %\addlegendentry{\scriptsize SDT48}
    \addplot[red!60,mark=otimes*,line width=1pt] coordinates {(1,0.276715934) (2,0.173591197) (5,0.164033324)
    (9,0.146789342) (13,0.138284281) (19,0.139886647) (25,0.142813057) (31,0.144191474) (37,0.137800246) (43,0.148809344) (48,0.152258441)};
    \addlegendentry{\scriptsize Base48}
    \end{axis}
    }
  \end{tikzpicture}
  \caption{The comparison of Emb Similarity between Base48 and \textsc{Sdt}-48L systems.}\label{fig:comparison}
\end{figure}
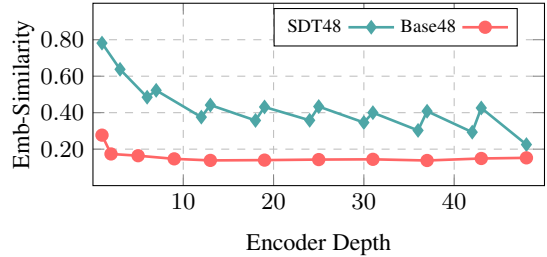

\begin{table}[!t]
  \small
  \setlength{\tabcolsep}{4.5pt}
  \begin{center}
  \begin{tabular}{llr}
  \toprule
  \textbf{System} & \textbf{BLEU} & \textbf{$\bigtriangleup$\bf BLEU}\\
  \midrule
  Pre-Norm-6L         & 27.05     &0\\
  Pre-Norm-12L        & 28.33     &$\Uparrow$ 1.28\\          
  Stack-12L           & 28.04     &$\Uparrow$ 0.99\\
  Reg-6L              & 27.45     &$\Uparrow$ 0.40\\
  %\hdashline
  %Base12        & asds  \\
  %Base24        & 28.95 \\
  %Double24      & 28.07 \\
  %\hline
  \bottomrule
  \end{tabular}
  \end{center}
  \caption{BLEU scores [$\%$] of several systems.}\label{tab:exp}
\end{table}

\subsection{Similarity of Layers}

We show that the inter-similarity and emb-similarity of shallow models fail to converge in Section \ref{sec:why}. Here, we further study the model behavior for different systems. We fix the parameters of a well-trained Pre-Norm-6L baseline and stack it into a $12$-layer system by copying the top-6 layers. The dash lines in Figure \ref{fig:sim-stack} denote the new stacked system Stack-12L. We observe that both inter-similarity and emb-similarity continue to rise and decline respectively, which exhibit similar phenomenon with Pre-Norm-12L which is trained from scratch. And it also outperforms the baseline by 0.99 BLEU points and is slightly inferior to Pre-Norm-12L (Table \ref{tab:exp}). Motivated by this, we design a regularization (according to Inter-sim) to constrain each layer to learn more global information on Pre-Norm-6L. Experimental results show that Reg-6L outperforms the baseline by 0.4 BLEU scores, indicating that a stronger global representation substantially improves the NMT model. This confirms our hypothesis in Section \ref{sec:why}.

Figure \ref{fig:comparison} plots the emb-similarity of Base48 and \textsc{Sdt}-48L. The model trained from shallow to deep behaves similarly with learning from scratch. The emb-similarity shows the same trend of decreasing in terms of similarity, and tends to coverage after layer 48. The results also indicate that our method can enbale the deep models to learn efficiently.

\section{Related Work}
\label{sec:related-work}

In this section, we discuss the related work from two aspects as follows:

\subsection{Deep Network Modeling}
In recent years, researchers gradually concentrate on building deep networks for Transformer \cite{vaswani2017attention}.
\citet{pham2019very} developed a 48-layer Transformer for speech recognition and adopted the stochastic residual connection to alleviate gradient vanishing/exploding problem.  \citet{bapna-etal-2018-training} demonstrated the challenge when training deep encoder models with vanilla Transformer on NMT task, due to the gradient vanishing or exploding. They also proposed a transparent attention mechanism to alleviate the problem. \citet{wang-etal-2019-learning} demonstrated the essential of layer-normalization in each layer and proposed the dynamic linear combination method to ease the information flow. Homochronously, \cite{wu-etal-2019-depth} trained a 8-layer Transformer-Big with three specially designed components. More recently, \citet{wei2004multiscale} further enhanced the Transformer-Big up to 18 layers through a multiscale collaborative framework. In general, shortening the path from bottom to top can obtain consistent improvements in the aforementioned studies. On the other hand, researchers observed that proper initialization strategies without any structure adjustment can also ease the optimization of Post-Norm Transformer, which highlighted the importance of careful parameter-initialization \cite{zhang-etal-2019-improving, xu2019lipschitz, Huang2020improving}.

\subsection{Efficient Training Methods}
When the model goes deeper, a challenge is the long training time for model convergence and the huge GPU cost. To alleviate this issue, several attempts have been made. \citet{ChangMHTB18} proposed a multi-level training method by interpolating a residual block right after each existing block to accelerate the training of ResNets in computer version. Similarly, \citet{GongHLQWL19} adopted a progressive stacking strategy to transfer the knowledge from a shallow model to a deep model, thus successfully trained a large-scale pre-training model BERT \cite{devlin-etal-2019-bert} at a faster rate with comparable performance on downstream tasks. Unlike previous work, we only copy parameters of the $g$ top-most layers and employ sparse connections across each stacking block in our shallow to deep training method, which has not been discussed yet in learning deep MT models.

\section{Conclusions}
\label{sec:conclusion}

We have investigated the behaviour of the well-trained deep Transformer models and found that stacking more layers could improve the representation ability of NMT systems. Higher layers share more global information over different positions and adjacent layers behave similarly. Also, we have developed a shallow-to-deep training strategy and employ sparse connections across blocks to ease the optimization. With the help of learning rate restart and appropriate initialization we successfully train a 48-layer RPR model by progressive stacking and achieve a $40\%$ speedup on both WMT'16 English-German and WMT'14 English-French tasks. Furthermore, our \textsc{Sdt}-RPR-24L (Big) achieves a BLEU score of $30.46$ on WMT'16 English-German task, and speeds up the training by $1.5\times$.

\section*{Acknowledgments}
This work was supported in part by the National Science Foundation of China (Nos. 61876035 and 61732005), the National Key R\&D Program of China (No. 2019QY1801). The authors would like to thank anonymous reviewers for their valuable comments. And thank Qiang Wang for the helpful advice to improve the paper.

\bibliography{emnlp2020}

\begin{thebibliography}{29}
\expandafter\ifx\csname natexlab\endcsname\relax\def\natexlab#1{#1}\fi

\bibitem[{Bahdanau et~al.(2015)Bahdanau, Cho, and Bengio}]{bahdanau2014neural}
Dzmitry Bahdanau, Kyunghyun Cho, and Yoshua Bengio. 2015.
\newblock Neural machine translation by jointly learning to align and
  translate.
\newblock In \emph{In Proceedings of the 3rd International Conference on
  Learning Representations}.

\bibitem[{Bapna et~al.(2018)Bapna, Chen, Firat, Cao, and
  Wu}]{bapna-etal-2018-training}
Ankur Bapna, Mia Chen, Orhan Firat, Yuan Cao, and Yonghui Wu. 2018.
\newblock Training deeper neural machine translation models with transparent
  attention.
\newblock In \emph{Proceedings of the 2018 Conference on Empirical Methods in
  Natural Language Processing}, pages 3028--3033, Brussels, Belgium.
  Association for Computational Linguistics.

\bibitem[{Chang et~al.(2018)Chang, Meng, Haber, Tung, and Begert}]{ChangMHTB18}
Bo~Chang, Lili Meng, Eldad Haber, Frederick Tung, and David Begert. 2018.
\newblock Multi-level residual networks from dynamical systems view.
\newblock In \emph{6th International Conference on Learning Representations,
  {ICLR} 2018, Vancouver, BC, Canada, April 30 - May 3, 2018, Conference Track
  Proceedings}.

\bibitem[{Devlin et~al.(2019)Devlin, Chang, Lee, and
  Toutanova}]{devlin-etal-2019-bert}
Jacob Devlin, Ming-Wei Chang, Kenton Lee, and Kristina Toutanova. 2019.
\newblock {BERT}: Pre-training of deep bidirectional transformers for language
  understanding.
\newblock In \emph{Proceedings of the 2019 Conference of the North {A}merican
  Chapter of the Association for Computational Linguistics: Human Language
  Technologies, Volume 1 (Long and Short Papers)}, pages 4171--4186,
  Minneapolis, Minnesota. Association for Computational Linguistics.

\bibitem[{Gehring et~al.(2017)Gehring, Auli, Grangier, Yarats, and
  Dauphin}]{gehring2017convs2s}
Jonas Gehring, Michael Auli, David Grangier, Denis Yarats, and Yann Dauphin.
  2017.
\newblock Convolutional sequence to sequence learning.
\newblock In \emph{ICML}.

\bibitem[{Gong et~al.(2019)Gong, He, Li, Qin, Wang, and Liu}]{GongHLQWL19}
Linyuan Gong, Di~He, Zhuohan Li, Tao Qin, Liwei Wang, and Tie{-}Yan Liu. 2019.
\newblock Efficient training of {BERT} by progressively stacking.
\newblock In \emph{Proceedings of the 36th International Conference on Machine
  Learning, {ICML} 2019, 9-15 June 2019, Long Beach, California, {USA}}, pages
  2337--2346.

\bibitem[{Greff et~al.(2017)Greff, Srivastava, and Schmidhuber}]{iclrGreff2017}
Klaus Greff, Rupesh~Kumar Srivastava, and J{\"{u}}rgen Schmidhuber. 2017.
\newblock Highway and residual networks learn unrolled iterative estimation.
\newblock In \emph{5th International Conference on Learning Representations,
  {ICLR} 2017, Toulon, France, April 24-26, 2017, Conference Track
  Proceedings}.

\bibitem[{He et~al.(2016)He, Zhang, Ren, and Sun}]{he2016deep}
Kaiming He, Xiangyu Zhang, Shaoqing Ren, and Jian Sun. 2016.
\newblock Deep residual learning for image recognition.
\newblock In \emph{Proceedings of the IEEE conference on computer vision and
  pattern recognition}, pages 770--778.

\bibitem[{Huang et~al.(2020)Huang, Perez, Ba, and Volkovs}]{Huang2020improving}
Xiao~Shi Huang, Felipe Perez, Jimmy Ba, and Maksims Volkovs. 2020.
\newblock Improving transformer optimization through better initialization.
\newblock In \emph{Proceedings of Machine Learning and Systems 2020}, pages
  9868--9876.

\bibitem[{Kingma and Ba(2015)}]{kingma2014adam}
Diederik~P. Kingma and Jimmy Ba. 2015.
\newblock Adam: {A} method for stochastic optimization.
\newblock In \emph{3rd International Conference on Learning Representations,
  {ICLR} 2015, San Diego, CA, USA, May 7-9, 2015, Conference Track
  Proceedings}.

\bibitem[{Lei~Ba et~al.(2016)Lei~Ba, Kiros, and Hinton}]{lei2016layer}
Jimmy Lei~Ba, Jamie~Ryan Kiros, and Geoffrey~E Hinton. 2016.
\newblock Layer normalization.
\newblock \emph{arXiv preprint arXiv:1607.06450}.

\bibitem[{Li et~al.(2019)Li, Li, Xu, Lin, Liu, Liu, Wang, Zhang, Xu, Wang,
  Feng, Chen, Liu, Li, Wang, Xiao, and Zhu}]{li-etal-2019-niutrans}
Bei Li, Yinqiao Li, Chen Xu, Ye~Lin, Jiqiang Liu, Hui Liu, Ziyang Wang, Yuhao
  Zhang, Nuo Xu, Zeyang Wang, Kai Feng, Hexuan Chen, Tengbo Liu, Yanyang Li,
  Qiang Wang, Tong Xiao, and Jingbo Zhu. 2019.
\newblock The {N}iu{T}rans machine translation systems for {WMT}19.
\newblock In \emph{Proceedings of the Fourth Conference on Machine Translation
  (Volume 2: Shared Task Papers, Day 1)}, pages 257--266, Florence, Italy.
  Association for Computational Linguistics.

\bibitem[{Liu et~al.(2020)Liu, Liu, Gao, Chen, and Han}]{liu2020understanding}
Liyuan Liu, Xiaodong Liu, Jianfeng Gao, Weizhu Chen, and Jiawei Han. 2020.
\newblock Understanding the difficulty of training transformers.
\newblock \emph{arXiv preprint arXiv:2004.08249}.

\bibitem[{Ott et~al.(2019)Ott, Edunov, Baevski, Fan, Gross, Ng, Grangier, and
  Auli}]{ott-etal-2019-fairseq}
Myle Ott, Sergey Edunov, Alexei Baevski, Angela Fan, Sam Gross, Nathan Ng,
  David Grangier, and Michael Auli. 2019.
\newblock fairseq: A fast, extensible toolkit for sequence modeling.
\newblock In \emph{Proceedings of the 2019 Conference of the North {A}merican
  Chapter of the Association for Computational Linguistics (Demonstrations)},
  pages 48--53, Minneapolis, Minnesota. Association for Computational
  Linguistics.

\bibitem[{Ott et~al.(2018)Ott, Edunov, Grangier, and Auli}]{ott-EtAl:2018:WMT}
Myle Ott, Sergey Edunov, David Grangier, and Michael Auli. 2018.
\newblock Scaling neural machine translation.
\newblock In \emph{Proceedings of the Third Conference on Machine Translation,
  Volume 1: Research Papers}, pages 1--9, Belgium, Brussels. Association for
  Computational Linguistics.

\bibitem[{Pham et~al.(2019)Pham, Nguyen, Niehues, M{\"{u}}ller, and
  Waibel}]{pham2019very}
Ngoc{-}Quan Pham, Thai{-}Son Nguyen, Jan Niehues, Markus M{\"{u}}ller, and Alex
  Waibel. 2019.
\newblock Very deep self-attention networks for end-to-end speech recognition.
\newblock In \emph{Interspeech 2019, 20th Annual Conference of the
  International Speech Communication Association, Graz, Austria, 15-19
  September 2019}, pages 66--70. {ISCA}.

\bibitem[{Sennrich et~al.(2016)Sennrich, Haddow, and
  Birch}]{sennrich-subword-neural}
Rico Sennrich, Barry Haddow, and Alexandra Birch. 2016.
\newblock Neural machine translation of rare words with subword units.
\newblock In \emph{Proceedings of the 54th Annual Meeting of the Association
  for Computational Linguistics (Volume 1: Long Papers)}, pages 1715--1725,
  Berlin, Germany. Association for Computational Linguistics.

\bibitem[{Shaw et~al.(2018)Shaw, Uszkoreit, and Vaswani}]{shaw-etal-2018-self}
Peter Shaw, Jakob Uszkoreit, and Ashish Vaswani. 2018.
\newblock Self-attention with relative position representations.
\newblock In \emph{Proceedings of the 2018 Conference of the North {A}merican
  Chapter of the Association for Computational Linguistics: Human Language
  Technologies, Volume 2 (Short Papers)}, pages 464--468, New Orleans,
  Louisiana. Association for Computational Linguistics.

\bibitem[{Sutskever et~al.(2014)Sutskever, Vinyals, and
  Le}]{sutskever2014sequence}
Ilya Sutskever, Oriol Vinyals, and Quoc~V Le. 2014.
\newblock Sequence to sequence learning with neural networks.
\newblock In \emph{Advances in neural information processing systems}, pages
  3104--3112.

\bibitem[{Vaswani et~al.(2017)Vaswani, Shazeer, Parmar, Uszkoreit, Jones,
  Gomez, Kaiser, and Polosukhin}]{vaswani2017attention}
Ashish Vaswani, Noam Shazeer, Niki Parmar, Jakob Uszkoreit, Llion Jones,
  Aidan~N Gomez, {\L}ukasz Kaiser, and Illia Polosukhin. 2017.
\newblock Attention is all you need.
\newblock In \emph{Advances in Neural Information Processing Systems}, pages
  6000--6010.

\bibitem[{Wang et~al.(2019)Wang, Li, Xiao, Zhu, Li, Wong, and
  Chao}]{wang-etal-2019-learning}
Qiang Wang, Bei Li, Tong Xiao, Jingbo Zhu, Changliang Li, Derek~F. Wong, and
  Lidia~S. Chao. 2019.
\newblock Learning deep transformer models for machine translation.
\newblock In \emph{Proceedings of the 57th Annual Meeting of the Association
  for Computational Linguistics}, pages 1810--1822, Florence, Italy.
  Association for Computational Linguistics.

\bibitem[{Wang et~al.(2018)Wang, Li, Xiao, Li, Li, and
  Zhu}]{wang-etal-2018-multi-layer}
Qiang Wang, Fuxue Li, Tong Xiao, Yanyang Li, Yinqiao Li, and Jingbo Zhu. 2018.
\newblock Multi-layer representation fusion for neural machine translation.
\newblock In \emph{Proceedings of the 27th International Conference on
  Computational Linguistics}, pages 3015--3026, Santa Fe, New Mexico, USA.
  Association for Computational Linguistics.

\bibitem[{Wei et~al.(2020)Wei, Yu, Hu, Zhang, Weng, and
  Luo}]{wei2004multiscale}
Xiangpeng Wei, Heng Yu, Yue Hu, Yue Zhang, Rongxiang Weng, and Weihua Luo.
  2020.
\newblock Multiscale collaborative deep models for neural machine translation.
\newblock In \emph{Proceedings of the 58th Annual Meeting of the Association
  for Computational Linguistics}, pages 414--426, Online. Association for
  Computational Linguistics.

\bibitem[{Wu et~al.(2019{\natexlab{a}})Wu, Fan, Baevski, Dauphin, and
  Auli}]{wu2019pay}
Felix Wu, Angela Fan, Alexei Baevski, Yann~N. Dauphin, and Michael Auli.
  2019{\natexlab{a}}.
\newblock Pay less attention with lightweight and dynamic convolutions.
\newblock In \emph{7th International Conference on Learning Representations,
  {ICLR} 2019, New Orleans, LA, USA, May 6-9, 2019}.

\bibitem[{Wu et~al.(2019{\natexlab{b}})Wu, Wang, Xia, Tian, Gao, Qin, Lai, and
  Liu}]{wu-etal-2019-depth}
Lijun Wu, Yiren Wang, Yingce Xia, Fei Tian, Fei Gao, Tao Qin, Jianhuang Lai,
  and Tie-Yan Liu. 2019{\natexlab{b}}.
\newblock Depth growing for neural machine translation.
\newblock In \emph{Proceedings of the 57th Annual Meeting of the Association
  for Computational Linguistics}, pages 5558--5563, Florence, Italy.
  Association for Computational Linguistics.

\bibitem[{Wu et~al.(2016)Wu, Schuster, Chen, Le, Norouzi, Macherey, Krikun,
  Cao, Gao, Macherey et~al.}]{wu2016google}
Yonghui Wu, Mike Schuster, Zhifeng Chen, Quoc~V Le, Mohammad Norouzi, Wolfgang
  Macherey, Maxim Krikun, Yuan Cao, Qin Gao, Klaus Macherey, et~al. 2016.
\newblock Google's neural machine translation system: Bridging the gap between
  human and machine translation.
\newblock \emph{arXiv preprint arXiv:1609.08144}.

\bibitem[{Xiao et~al.(2019)Xiao, Li, Zhu, Yu, and Liu}]{xiao-etal-2019-sharing}
Tong Xiao, Yinqiao Li, Jingbo Zhu, Zhengtao Yu, and Tongran Liu. 2019.
\newblock Sharing attention weights for fast transformer.
\newblock In \emph{Proceedings of the Twenty-Eighth International Joint
  Conference on Artificial Intelligence, {IJCAI} 2019, Macao, China, August
  10-16, 2019}, pages 5292--5298. ijcai.org.

\bibitem[{Xu et~al.(2020)Xu, Liu, van Genabith, Xiong, and
  Zhang}]{xu2019lipschitz}
Hongfei Xu, Qiuhui Liu, Josef van Genabith, Deyi Xiong, and Jingyi Zhang. 2020.
\newblock Lipschitz constrained parameter initialization for deep transformers.
\newblock In \emph{Proceedings of the 58th Annual Meeting of the Association
  for Computational Linguistics}, pages 397--402, Online. Association for
  Computational Linguistics.

\bibitem[{Zhang et~al.(2019)Zhang, Titov, and
  Sennrich}]{zhang-etal-2019-improving}
Biao Zhang, Ivan Titov, and Rico Sennrich. 2019.
\newblock Improving deep transformer with depth-scaled initialization and
  merged attention.
\newblock In \emph{Proceedings of the 2019 Conference on Empirical Methods in
  Natural Language Processing and the 9th International Joint Conference on
  Natural Language Processing (EMNLP-IJCNLP)}, pages 897--908, Hong Kong,
  China. Association for Computational Linguistics.

\end{thebibliography}
\bibliographystyle{acl_natbib}

\end{document}